\documentclass[10pt,twocolumn]{article}

\usepackage[margin=0.72in]{geometry}
\usepackage{graphicx}
\usepackage{amsmath,amssymb}
\usepackage{booktabs}
\usepackage{array}
\usepackage{tabularx}
\usepackage{tikz}
\usepackage{url}
\usepackage[hidelinks]{hyperref}
\usepackage[hypcap=false]{caption}
\usepackage{microtype}
\usepackage{cuted}
\usetikzlibrary{calc,fit,positioning}

\hypersetup{
  pdftitle={From Checker to Forecaster: Code-Owned Evaluation of Model-Generated Strategic Routes Under Delayed Ground Truth},
  pdfauthor={Aleh Manchuliantsau, Independent Researcher, aleh.manchuliantsau@gmail.com},
  pdfsubject={Code-owned provisional evaluation under delayed ground truth},
  pdfkeywords={LLM evaluation, delayed ground truth, strategic routes, calibration, decision analysis}
}

\title{\vspace{-1.2em}From Checker to Forecaster: Code-Owned Evaluation\\of Model-Generated Strategic Routes Under Delayed Ground Truth}
\author{Aleh Manchuliantsau\\Independent Researcher\\\texttt{aleh.manchuliantsau@gmail.com}}
\date{Version 1.0: July 2026}

\begin{document}
\maketitle

\begin{abstract}
Many evaluations of model outputs rely either on contracts checkable at evaluation time or on feedback that arrives within the operating loop. We study the complementary setting in which ground truth is delayed, censored, or private, so deterministic code cannot check correctness at scoring time and must instead issue a code-owned provisional forecast. RouteCast instantiates this regime for model-generated typed strategic routes: models propose candidate routes and structured factors; point-in-time evidence, reference classes, and deterministic transformations produce a provisional forecast-ranking; later outcomes evaluate the forecast. In a retrospective venture pilot on 21 binary-outcome cases (6 positive, 15 negative), the whole-packet RouteCast score showed preliminary retrospective discrimination (AUC 0.756, 95\% CI \([0.471,0.980]\)), while a blind LLM judge reached AUC 0.678 \([0.419,0.897]\) and an identity-exposed LLM judge reached AUC 0.761 \([0.515,0.944]\), consistent with recognition- or outcome-related leakage risk. A preregistered decomposition ablation on the same binary subset found that converting the identical inputs into typed staged routes was indistinguishable from the whole-packet score (\(\Delta\mathrm{AUC}=-0.144\), 95\% CI \([-0.471,0.176]\)) and from a deterministic heuristic (\(\Delta\mathrm{AUC}=-0.089\), 95\% CI \([-0.412,0.278]\)). The pilot establishes an auditable feasibility result and exposes failure modes; it does not establish prospective calibration, causal decision improvement, route-decomposition advantage, or cross-domain validity.
\end{abstract}

\section{Introduction}

Models increasingly generate strategic options rather than only factual answers. A founder asks which venture route to pursue, a lab asks which research direction might be worth a quarter, and a product team asks which roadmap bet should be made next. These tasks differ from factual question answering because the relevant outcome arrives later, is often censored or private, and may be observed only through a proxy. The evaluator must act before the world can grade the recommendation.

Deterministic evaluation of model outputs has so far required either a contract checkable at evaluation time or feedback arriving within the loop. We study the complementary regime---delayed, censored, or private ground truth---in which deterministic code must issue a code-owned provisional forecast whose correctness is unknowable at scoring time. We instantiate this regime for a new evaluation object, competing model-generated typed strategic routes, and audit the protocol's integrity properties on a retrospective pilot, including an identity-leakage contrast and a preregistered decomposition ablation reported as indistinguishable.

The conceptual move is from checker to forecaster. Deterministic code is already used as a checker or control authority in software evaluation, runtime assurance, and evidence-gated control. In the regime studied here, however, correctness is not available at scoring time. Code can own the forecast only by issuing an auditable forecast-like ranking, preserving the information set available at decision time, and later exposing that ranking to outcome resolution. The retrospective pilot therefore asks whether the implementation survives obvious integrity failures. It is not a validation of prospective forecasting authority.

\begin{table*}[t]
\centering
\small
\setlength{\tabcolsep}{4pt}
\caption{Feature comparison with representative close predecessors.}
\label{tab:regime-map}
\begin{tabularx}{\textwidth}{@{}>{\raggedright\arraybackslash}p{2.25cm}>{\raggedright\arraybackslash}p{2.35cm}>{\raggedright\arraybackslash}p{4.1cm}>{\raggedright\arraybackslash}p{3.25cm}>{\raggedright\arraybackslash}X@{}}
\toprule
System & Outcome resolution & Inputs to final rule & Final rule & Evaluation object \\
\midrule
GroundEval / CARE &
checkable now or in-loop &
observed traces or evidence gates &
deterministic check/gate &
agent behavior or policy \\
\addlinespace
DeLLMa &
delayed &
LLM verbal forecasts and LLM preference rankings &
analytic expected-utility maximization &
single action under uncertain states \\
\addlinespace
DIALECTIC &
delayed &
model scores or judgments &
model-owned or model-aggregated &
company, investment, or rationale \\
\addlinespace
RouteCast &
delayed, censored, or private &
frozen priors and admissibility-checked evidence, with model-mediated typing and extraction &
versioned deterministic forecast-ranking &
competing typed strategic routes \\
\bottomrule
\end{tabularx}
\end{table*}

The initial implementation domain is venture route selection. RouteCast ranks Wedge--Bridge--Vision trajectories from point-in-time packets. A route packet can contain a current wedge, typed transition claims, evidence, cost-to-learn, a staged-value calculation, and a binding transition to test. RouteCast is defined for competing typed strategic routes. The retrospective pilot evaluates one cohort through several analyses: a frozen whole-packet RouteCast score, blinded and identity-exposed LLM baselines, a deterministic heuristic, a bounded stability probe, and a preregistered typed-route decomposition ablation. The decomposition analysis is therefore a variant within the same pilot, not evidence from a separate pilot.

The contributions are:
\begin{enumerate}
\item \textbf{Regime formulation.} We formalize code-owned provisional evaluation when correctness cannot be checked at scoring time.
\item \textbf{Protocol and object.} We instantiate the regime for competing typed strategic routes using point-in-time evidence, reference classes, deterministic ranking, and a binding-transition output.
\item \textbf{Retrospective pilot.} We report an identity-exposure contrast, preliminary whole-packet discrimination, and a preregistered decomposition ablation reported as indistinguishable.
\end{enumerate}

The scope is deliberately bounded. We do not claim that RouteCast has an established advantage over LLM judges, that its probabilities are calibrated, that typed decomposition improved ranking in the pilot, or that the protocol transfers to OSS, R\&D, or product roadmaps. Those are prospective evaluation targets.

\section{Problem Setting and Definitions}

Let \(t_0\) denote the decision time. The available information set \(\mathcal{I}_{t_0}\) contains only evidence admissible at that time: captured source text, timestamps, provenance, reference classes, and user-provided context. The outcome \(Y\) is unavailable, censored, private, or unresolved at \(t_0\). At a later resolution time \(t_1\), a frozen predicate or outcome mapping may reveal \(Y\), a proxy for \(Y\), or an unresolved status.

\begin{center}
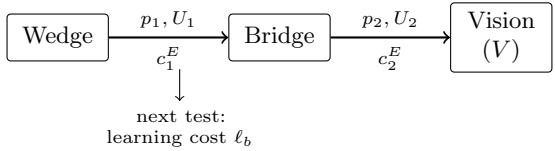

\vspace{0.4em}
\centering
\begin{tikzpicture}[
  node distance=1.58cm,
  state/.style={draw, rounded corners=1.4pt, align=center, minimum width=1.34cm, minimum height=0.62cm, font=\small},
  edge/.style={->, thick},
  note/.style={font=\scriptsize, align=center}
]
\node[state] (wedge) {Wedge};
\node[state, right=of wedge] (bridge) {Bridge};
\node[state, right=of bridge] (vision) {Vision\\(\(V\))};

\draw[edge] (wedge) -- node[above, font=\scriptsize] {\(p_1, U_1\)}
  node[below, font=\scriptsize] {\(c_1^E\)} (bridge);
\draw[edge] (bridge) -- node[above, font=\scriptsize] {\(p_2, U_2\)}
  node[below, font=\scriptsize] {\(c_2^E\)} (vision);

\draw[->, thin] ($(wedge)!0.55!(bridge)+(0,-0.42)$) -- +(0,-0.42)
  node[below, note, text width=2.2cm] {next test:\\learning cost \(\ell_b\)};
\end{tikzpicture}
\captionof{figure}{Typed strategic route. Each transition \(e_i\) carries a provisional success probability \(p_i\), epistemic uncertainty \(U_i\), and an execution cost \(c_i^E\). A selected binding transition can be tested at learning cost \(\ell_i\) before full execution. The terminal node carries value \(V\).}
\label{fig:typed-route}
\vspace{0.2em}
\end{center}

A \emph{strategic route} is a candidate path from a current state to a target state. A \emph{typed transition} is one edge in that path with a source state, target state, actor, mechanism, required asset, evidence, threshold, kill condition, probability, cost, and, when applicable, resolution window. A \emph{model-owned proposal} is any route, claim, decomposition, prior suggestion, or rationale authored by a model. A \emph{code-owned provisional forecast} is a ranking or derived quantity produced by versioned code and data from \(\mathcal{I}_{t_0}\), without permitting model prose to overwrite the final transformation.

Three evaluation concepts must remain separate. \emph{Ranking performance} asks whether later positives rank above later negatives, for example by AUC or top-bucket lift. \emph{Probability calibration} asks whether forecast probabilities match observed frequencies, for example by Brier score, calibration curves, or expected calibration error with small-sample caveats \cite{brier1950verification,gneiting2007proper,guo2017calibration}. \emph{Decision utility} asks whether acting on the ranking improves outcomes or reduces cost relative to a counterfactual policy. The retrospective pilot mainly provides evidence about integrity, retrospective discrimination, and failure modes. It does not measure prospective calibration or causal utility.

\section{Related Work}

\subsection{Proposal--Authority Separation and Runtime Assurance}

RouteCast does not introduce proposal--authority separation. The Simplex lineage and runtime-assurance literature separate a high-performance or untrusted controller from a trusted safety mechanism or fallback authority \cite{seto1998simplex,sha2001simplicity,fuller2020rta}. These works establish proposal--authority separation in settings where safety or correctness can be checked now, or where a trusted fallback/control mechanism operates in-loop.

Similar generator--verifier and diagnosis--control patterns appear in contemporary agent systems. For example, auditable policy-adaptation work separates an LLM diagnostic layer from a deterministic control layer \cite{auditablePolicyAdaptation2026}. The shared principle is that a model may propose, diagnose, or structure information without owning the final authority. RouteCast uses that principle in a setting where the final authority cannot be a present-time checker.

\subsection{Deterministic Computation Over Model-Derived Factors}

DeLLMa prompts an LLM for verbal state forecasts, converts them into a normalized state distribution, derives pairwise preferences from LLM rankings, fits an approximate utility function, and analytically maximizes expected utility \cite{liu2024dellma}. RouteCast is therefore not distinguished by deterministic arithmetic alone. Its residual distinction is the evaluation object and input-governance protocol: model outputs may provide typed classifications, extracted factors, evidence candidates, and estimator runs, but frozen priors, admissibility-checked evidence records, and versioned transformations produce the route quantities and forecast-ranking that are later evaluated against transition outcomes.

\subsection{Deterministic Checking and In-Loop Control}

GroundEval replaces LLM-as-judge scoring with deterministic, trace-based checks over stateful agent behavior, evidence access, and time-bounded constraints \cite{flynt2026groundeval}. It works because the evaluation contract is checkable now: the evaluator can inspect what the agent fetched, cited, and was allowed to access. CARE keeps a non-LLM incumbent optimizer as the default path in scientific experimentation and uses an auditable evidence gate before authorizing a challenger policy \cite{liu2026care}. It works because feedback arrives within the adaptation loop. RouteCast studies the complementary regime where correctness is unavailable at scoring time and the code-owned output must be graded later.

\subsection{Delayed-Outcome and Venture-Evaluation Systems}

Several recent systems evaluate investments, ventures, or finance rationales under delayed outcomes. DIALECTIC uses multi-agent debate and numerical scores for startup evaluation, with backtesting on opportunities from five VC funds \cite{bae2026dialectic}. ValueBlindBench frames delayed-ground-truth financial-rationale evaluation as a pre-calibration metrology problem for LLM judges \cite{chang2026valueblindbench}. Strategic Foresight reports a fully prospective venture tournament over Kickstarter projects \cite{csaszar2026strategic}. VCBench provides anonymized founder profiles for venture forecasting and explicitly addresses identity leakage \cite{chen2025vcbench}. Look-Ahead-Bench studies look-ahead bias in point-in-time financial LLM workflows \cite{benhenda2026lookahead}. SSFF combines machine-learning and LLM components for startup success forecasting \cite{wang2024ssff}.

These works motivate the delayed-ground-truth setting and provide important comparison points. They generally evaluate companies, founders, rationales, or single forecasts, often with model-owned scoring or model-derived or model-aggregated factors. RouteCast's residual target is narrower: code-owned provisional ranking of competing typed strategic routes under delayed, censored, or private ground truth.

\subsection{Residual Relationship to Prior Work}

RouteCast does not introduce deterministic decision arithmetic or delayed-outcome evaluation individually. Its residual contribution is their combination for competing typed strategic routes, with point-in-time provenance, frozen priors, admissibility-checked evidence, staged costs and values, and transition-level later resolution. The decision-analysis components draw on expected utility and statistical decision theory \cite{savage1954foundations,raiffa1961applied}, reference-class and outside-view reasoning \cite{kahneman1993timid}, staged investment and real-options logic \cite{cooper1990stagegate,dixit1994investment}, and value-of-information analysis \cite{howard1966information}.

\section{RouteCast Protocol}

RouteCast is a protocol for evidence-gated, code-owned provisional forecasting. The protocol has an authority boundary: models may propose route packets, assumptions, priors, and decompositions; evidence pipelines may produce admissible observations; versioned code and data own deterministic transformations and the forecast-ranking; later outcome resolution evaluates the forecast.

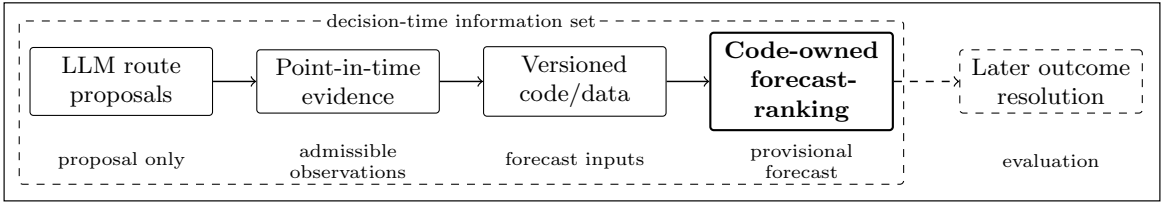
\begin{figure*}[t]
\centering
\begin{tikzpicture}[
  box/.style={draw, rounded corners=1.3pt, align=center, minimum height=0.78cm, text width=2.18cm, font=\small},
  future/.style={draw, dashed, rounded corners=1.3pt, align=center, minimum height=0.78cm, text width=2.18cm, font=\small},
  authority/.style={draw, line width=0.8pt, rounded corners=1.3pt, align=center, minimum height=0.88cm, text width=2.18cm, font=\small\bfseries},
  arrow/.style={->, line width=0.55pt},
  note/.style={align=center, font=\scriptsize, text width=2.35cm}
]
\node[box] (proposal) at (0,0) {LLM route\\proposals};
\node[box] (evidence) at (3.0,0) {Point-in-time\\evidence};
\node[box] (inputs) at (6.0,0) {Versioned\\code/data};
\node[authority] (forecast) at (9.0,0) {Code-owned\\forecast-ranking};
\node[future] (resolution) at (12.3,0) {Later outcome\\resolution};

\draw[arrow] (proposal) -- (evidence);
\draw[arrow] (evidence) -- (inputs);
\draw[arrow] (inputs) -- (forecast);
\draw[arrow, dashed] (forecast) -- (resolution);

\node[note] at (0,-1.05) {proposal only};
\node[note] at (3.0,-1.05) {admissible observations};
\node[note] at (6.0,-1.05) {forecast inputs};
\node[note] at (9.0,-1.05) {provisional forecast};
\node[note] at (12.3,-1.05) {evaluation};

\draw[dashed, rounded corners=1.4pt] (-1.35,0.80) rectangle (10.35,-1.36);
\node[font=\scriptsize, fill=white, inner sep=1pt] at (4.5,0.80) {decision-time information set};
\draw[line width=0.45pt] (-1.55,1.04) rectangle (13.74,-1.58);
\end{tikzpicture}
\caption{RouteCast authority-boundary architecture. Models propose routes and factors; point-in-time evidence, versioned code, and frozen data produce a code-owned provisional forecast-ranking that is evaluated only after later outcome resolution.}
\label{fig:routecast-architecture}
\end{figure*}

Code-owned does not mean objectively correct. It means the final transformation is explicit, reproducible, and not overwritten by model prose. A code-owned forecast-ranking can be wrong, miscalibrated, or overfit. The point is that the ranking can be audited, ablated, frozen, and later scored.

Code-owned also does not mean model-free. Models may classify transitions, extract factors, propose evidence records, or contribute estimator runs. These outputs affect the computation only through typed and provenance-checked interfaces. The ownership claim is narrower: final numerical quantities and rankings are produced by a frozen deterministic pipeline, and model prose cannot directly overwrite them.

Operationally, the protocol uses packetized provenance. A proposal packet records candidate routes and claims. An evidence packet records source excerpts, capture dates, trust tiers, corroborating and contradicting signals, and the point-in-time boundary. A ranking packet records code-computed quantities such as normalized scores, probabilities, costs, values, risk gates, and the selected binding transition. This packet structure makes leakage checks and future resolution possible.

\section{Route Object and Forecasting}

A typed route is a sequence or graph of transitions:
\[
e_i=(s_i,s_{i+1},a_i,m_i,A_i,E_i,\theta_i,k_i,p_i,U_i,c_i^E,\ell_i,h_i),
\]
where \(s_i\) and \(s_{i+1}\) are source and target states, \(a_i\) is the actor, \(m_i\) the mechanism, \(A_i\) the required asset, \(E_i\) the evidence set, \(\theta_i\) the success threshold, \(k_i\) the kill condition, \(p_i\) the provisional transition probability, \(U_i\) epistemic uncertainty, \(c_i^E\) execution or continuation cost, \(\ell_i\) learning cost, and \(h_i\) a resolution horizon when one is specified.

\(p_i\) is the provisional probability that transition \(e_i\) will resolve positively under the information set available at decision time. \(U_i\) measures epistemic uncertainty around that estimate, for example disagreement or sensitivity across admissible estimators. It is not the transition's failure probability \(1-p_i\). In the route object, \(p_i\) enters route-value forecasting, while \(U_i\) helps identify which uncertainty should be tested next. \(U_i\) is not automatically subtracted from \(p_i\), and the present uncertainty measure is a heuristic rather than a calibrated confidence interval. When an implementation uses cross-family or cross-model disagreement, that disagreement is an implementation heuristic rather than a calibrated uncertainty estimator.

The descriptors ``code-computed'' and ``code-owned'' refer to the deterministic transformation and final authority, not to model-independent provenance. The final \(p_i\) is code-derived but may be model-mediated through transition typing and evidence extraction. \(U_i\) is a code-computed proxy derived partly from model-estimator disagreement.

\subsection{Code-Owned Arithmetic and Gates}

Models may produce per-criterion rationales and structured sub-scores, but the aggregate arithmetic is code-owned. Criterion totals are computed as
\[
T =
\frac{\sum_i w_i \cdot \operatorname{clip}_{[0,100]}(s_i)}
{\sum_i w_i},
\]
over present criteria. Missing criteria are excluded from the denominator; model-stated totals are ignored.

Eligibility and risk gates are applied before ranking and cannot be averaged away by a high narrative or value score; their definitions and thresholds are versioned and frozen before evaluation.

\subsection{Evidence Reweighting}

The implemented protocol adjusts a prior from corroborating and contradicting evidence tiers:
\[
p(e)=\operatorname{clip}_{[0,1]}
\left(p_{\mathrm{prior}}(e)\cdot W_{\mathrm{reality}}(e)\right),
\]
\[
W_{\mathrm{reality}}(e)=
\operatorname{clip}_{[0.3,1.5]}
\left(1+0.5\operatorname{sat}(C)-0.5\operatorname{sat}(K)\right),
\]
where \(C\) and \(K\) are corroborating and contradicting evidence weights and
\[
\operatorname{sat}(x)=\frac{\sum_i x_{(i)}0.5^i}{1+\sum_i x_{(i)}0.5^i}.
\]
\(p_{\mathrm{prior}}(e)\) is drawn from a frozen reference-class table or another versioned prior policy; it is not accepted directly from model prose. Where no calibrated prior exists, the value remains a preregistered proxy rather than a calibrated probability claim. In the saturation function, \(x_{(i)}\) denotes evidence weights sorted in descending order. If no admissible evidence is available, \(W_{\mathrm{reality}}=1\). Unverified evidence is excluded before computation. The rule is code-owned and saturating, so additional evidence has diminishing marginal effect. Its marginal contribution is not isolated by the retrospective pilot.

\subsection{Staged Value, Execution Cost, and Cost-to-Learn}

Execution costs are computed bottom-up and floored by reference classes:
\[
c_i^E=\max\left(\sum_j n_{ij} u_{ij},\; c_{\mathrm{reference\ floor}}(e_i)\right).
\]
Route value is folded as staged expected value:
\[
EV=-c_1^E+p_1\left(-c_2^E+p_2\left(\cdots+p_nV\right)\right).
\]
The terminal value proxy used by the pilot is
\[
V=\mathrm{TAM}\times \alpha_{\mathrm{capture}}\times \alpha_{\mathrm{defensibility}}.
\]
This is a frozen ranking proxy, not an empirically calibrated company valuation. Its dominance in the decomposition ablation is reported as a construct-validity limitation. The staged-value formula is an implementation of standard staged decision and real-options reasoning, not a claim of mathematical novelty. The semantics are abandonment-aware: later execution costs are paid only if earlier transitions succeed.

The learning cost \(\ell_i\) is different from \(c_i^E\). It is the cost of an experiment or evidence-gathering action intended to reduce \(U_i\) or resolve the transition predicate before full execution. \(\ell_i\) does not automatically enter the existing frozen expected-value formula unless the learning experiment itself is modeled as a route stage.

\subsection{Binding Transition}

The binding transition is a VoI-inspired heuristic for the next uncertainty to test:
\[
b=\arg\max_i U_iD_i.
\]
Here \(U_i\) is uncertainty and \(D_i\) is downstream stake. The current heuristic identifies a high-leverage uncertainty and reports its learning cost separately. A cost-aware value-of-information or knowledge-gradient selector belongs in prospective evaluation. The heuristic is a protocol output, not established decision utility.

\subsection{Flat Probability Product}

The flat product \(\prod_i p_i\) is retained only as a diagnostic because it ignores staged abandonment and is mechanically sensitive to route length.

\section{Retrospective As-Of Pilot}

The pilot serves as an integrity audit and feasibility test, not as a confirmatory backtest.

\subsection{Cohort and Outcome Mapping}

The retrospective pilot uses YC 2012--2014 companies with point-in-time descriptions reconstructed from public materials. The audit cohort contains 26 cases. Five cases with mixed outcome labels are retained in the case-level appendix but excluded from binary discrimination metrics. All reported AUC and lift estimates therefore use the 21-case binary subset: 6 positive and 15 negative.

The sample is subsampled and outcome-balanced rather than prevalence-representative. It is also not cleanly held out from prior development, so the estimates should be treated as integrity and feasibility evidence rather than confirmatory performance. Heterogeneous later outcomes were mapped separately before binary aggregation.

\subsection{Reconstruction and Blinding}

Point-in-time packets use descriptions and evidence intended to reflect what was admissible at the decision time, with Wayback verification where applicable. Later outcomes were stored separately from scoring inputs. For blinded packets, company names, domains, URLs, investor names, fame flags, and obvious brand cues were removed. Recognizability may remain from distinctive facts, and outcome raters may know later histories.

The decomposition run used masked point-in-time inputs. Network access was disabled and regression-tested for that run, and each company was scored once under the preregistered rule.

\subsection{Frozen Artifacts and Pilot Analyses}

The retrospective pilot contains a frozen whole-packet RouteCast score, a blind LLM judge with \(K=5\) correlated same-model passes, an identity-exposed LLM contrast, a simple deterministic heuristic, and a typed-route decomposition ablation. Score files were frozen before outcome join. The original whole-packet scores were frozen on 2026-07-07 before outcome labeling, with score-file SHA-256 beginning \texttt{4cc4dff}; the decomposition-ablation scores were frozen before outcome join at commit \texttt{b0dbd8a}, with SHA-256 beginning \texttt{de4172cd}. Both provisional score artifacts therefore predate this manuscript's arXiv/public-disclosure package. The repository preserves the legacy field name \texttt{crf\_score} in the frozen score artifact.

The pilot asks four questions. Does the whole-packet code-owned score retain retrospective discrimination? Does identity exposure change apparent LLM-judge discrimination? Does typed decomposition add discrimination? What protocol weaknesses are exposed by the retrospective pilot?

\section{Results}

\begin{table*}[t]
\centering
\small
\setlength{\tabcolsep}{5pt}
\caption{Retrospective pilot results on the 21-case binary subset. Raw scores are not cross-scorer comparable.}
\label{tab:pilot-audit-summary}
\begin{tabularx}{\textwidth}{@{}p{4.35cm}p{0.55cm}p{0.75cm}p{1.85cm}p{1.05cm}X@{}}
\toprule
Scorer & \(N\) & AUC \(\uparrow\) & 95\% CI & Top-Q lift \(\uparrow\) & Interpretation \\
\midrule
RouteCast whole-packet frozen score & 21 & 0.756 & [0.471, 0.980] & 2.1 & preliminary retrospective discrimination \\
Blind LLM judge mean \(K=5\) & 21 & 0.678 & [0.419, 0.897] & 1.4 & blinded LLM baseline \\
Simple deterministic heuristic & 21 & 0.700 & [0.406, 0.948] & 2.1 & deterministic baseline \\
Identity-exposed LLM judge & 21 & 0.761 & [0.515, 0.944] & 1.4 & leakage contrast only \\
\midrule
\multicolumn{6}{@{}p{\textwidth}@{}}{\emph{Note.} Single-scorer CIs are company bootstraps with \(B=10{,}000\) and seed 20260710. The table reports discrimination, not calibration. Across the five correlated passes, blind-LLM AUC ranged from 0.650 to 0.706; RouteCast re-extraction remained at 0.756 in all five runs. The legacy frozen score field is \texttt{crf\_score}.} \\
\bottomrule
\end{tabularx}
\end{table*}

\subsection{Whole-Packet Analysis}

The whole-packet RouteCast score showed retrospective discrimination on the 21-case binary analysis subset: AUC 0.756, with a 95\% company-bootstrap CI of \([0.471,0.980]\), with top-quartile lift 2.1. The sample is too small for general claims, and raw scores are not comparable across scoring systems; AUC and lift use within-scorer rankings.

\subsection{Identity-Exposure Contrast}

The blind LLM judge mean over \(K=5\) same-model passes reached AUC 0.678, with a 95\% company-bootstrap CI of \([0.419,0.897]\). The identity-exposed LLM judge reached AUC 0.761 \([0.515,0.944]\). Identity exposure increased apparent discrimination in this experimental contrast. The result is consistent with recognition- or outcome-related leakage, but it does not establish that leakage caused the improvement. Distinctive masked facts may still permit recognition, so the blind result is also not leakage-free.

The whole-packet RouteCast score and the exposed judge are not established as statistically different by this audit. A paired analysis would be needed for that claim, and the present paper does not include one. Retrospective discrimination is also not calibration.

\section{Decomposition Ablation}

The preregistered decomposition ablation tested whether converting the same masked point-in-time packets into typed staged routes improved cohort ranking. The experiment was preregistered before outcome join at commit \texttt{cb7625d}. Scores were frozen before outcome join at commit \texttt{b0dbd8a} in \texttt{eval/decomposed\_backtest/scores\_frozen.jsonl}, with SHA-256 beginning \texttt{de4172cd}. Network access was disabled and regression-tested, each company was run once, and the reporting rule was fixed in advance: a tie or loss would be reported.

\begin{table}[t]
\centering
\small
\caption{Decomposition ablation on the 21-case binary subset. Paired company bootstrap, \(B=10{,}000\), seed 20260710.}
\label{tab:decomposition-ablation}
\begin{tabularx}{\columnwidth}{Xcc}
\toprule
Comparison & \(\Delta\)AUC & 95\% CI \\
\midrule
Decomposed staged-NPV minus whole-packet RouteCast \texttt{crf\_score} & -0.144 & [-0.471, 0.176] \\
Decomposed staged-NPV minus deterministic heuristic & -0.089 & [-0.412, 0.278] \\
\midrule
Diagnostic: isolated chain probability product AUC & 0.333 & [0.092, 0.606] \\
\bottomrule
\end{tabularx}
\end{table}

Against the whole-packet RouteCast score,
\[
\Delta \mathrm{AUC}=-0.144,\quad 95\%\,\mathrm{CI}=[-0.471,0.176].
\]
Against the deterministic heuristic,
\[
\Delta \mathrm{AUC}=-0.089,\quad 95\%\,\mathrm{CI}=[-0.412,0.278].
\]
Both intervals span zero. The verdict under the preregistered rule is: indistinguishable on this cohort. The experiment provides no evidence that decomposition improves discrimination beyond the whole-packet protocol on this cohort.

The diagnostics are more useful than the point estimate. The isolated chain probability product had AUC 0.333 with 95\% CI \([0.092,0.606]\). Because it ignores staged abandonment and is mechanically sensitive to route length, it can anti-rank later positives when ambitious chains contain more transitions. Staged-NPV discrimination was dominated by terminal value/TAM rather than by the transition probability chain. The first edge was labeled observed in 58\% of generated chains, which suggests E0 inflation when a thin product description is treated as demonstrated traction. These diagnostics motivated protocol hardening and prospective evaluation rules; they are not evidence of calibrated edge forecasting. The null ablation limits the discrimination claim; it does not remove the value of typed transitions as preregistered resolution interfaces and tools for error localization.

\section{What the Evidence Supports}

The pilot supports an identity-exposure warning and provides preliminary evidence of whole-packet retrospective discrimination on this cohort. It does not support a typed-decomposition advantage. It does not establish calibration, incremental value from evidence reweighting, or decision utility. Cross-domain transfer and scorer self-improvement remain untested.

Discrimination and calibration are not interchangeable. AUC 0.756 \([0.471,0.980]\) on a small retrospective cohort does not imply that the probability assigned to a transition is numerically calibrated, and the negative decomposition ablation does not imply that typed routes are useless. It only says they did not add ranking signal in this pilot implementation on this cohort.

\section{Limitations}

\paragraph{Internal Validity.}
The retrospective pilot uses historical reconstruction. Packet construction involved human choices, outcome labels may compress heterogeneous trajectories, and point-in-time reconstruction can be contaminated by hindsight. Outcome raters may know later company histories. Masked facts can remain recognizable. Evidence-source selection and reference-class construction may also carry leakage or selection bias.

\paragraph{Statistical Validity.}
The binary cohort has \(n=21\) cases with 6 positives. AUC estimates are unstable and confidence intervals are wide. Multiple comparisons are present, and paired uncertainty is not resolved for every contrast. There is no independent external test cohort. Future transition-level analyses must cluster by route because edges within a route are not independent samples.

\paragraph{Construct Validity.}
Funding, traction, or persistence is not identical to route quality. Outcome labels compress pivots, acqui-hires, shutdowns, revenue, customer proof, and capital access. The decomposition ablation used one-sentence packets; its route chains are inferential reconstructions, not verified accounts of realized company plans. Terminal value/TAM dominance means staged NPV can recover a large-market signal rather than transition-level edge quality.

\paragraph{External Validity.}
The pilot is venture-only, historical, public-data-limited, and affected by survivorship and censored failures. It does not show transfer to OSS, R\&D, product roadmaps, or private operational settings. Market regimes and funding environments also change.

\paragraph{System Validity.}
Evidence acquisition remains partly manual. Code-owned output can be wrong, priors can be misspecified, and private outcomes may never resolve. Evidence reweighting, binding-transition selection, and route-level probabilities require prospective calibration before they can be treated as forecast-quality claims. Model-mediated transition typing may select an inappropriately favorable reference class. Cross-family agreement may also understate uncertainty when model families share correlated blind spots. Prospective evaluation should therefore include transition-type sensitivity, source deduplication, reference-class coverage diagnostics, and non-model uncertainty baselines.

\section{Prospective Protocol}

The Open Experiment is the planned implementation of the scientific protocol, not evidence that the retrospective pilot is calibrated. Cases should be registered prospectively, with frozen model and scorer versions, typed transitions created before outcomes, milestone predicates specified before resolution, and 30/90/180-day or domain-appropriate windows. Outcome resolution should be blinded where feasible.

The protocol should evaluate both discrimination and calibration. Metrics include Brier score, calibration curves, expected calibration error with binning caveats, AUC, top-bucket lift, and decision-relevant utility measures where a defensible counterfactual exists. Confidence intervals for transition-level metrics should use a cluster bootstrap by route.

Baselines should be credible enough to fail the method: a fitted logistic or gradient-boosted model on the same structured factors; a GroundEval-like contract baseline that checks evidence compliance; a CARE-like incumbent/challenger gate; a standard decision tree or flat-NPV baseline; a route-structure ablation; a reality-reweighting ablation; and a binding-transition comparison against a defensible value-of-information or knowledge-gradient baseline. Recognition should be measured as a covariate rather than treated as a binary nuisance. Prospective forecast calibration is the target of this program.

\section{Discussion}

\subsection{From Checker to Forecaster}

A checker evaluates against an available contract. A forecaster issues a provisional forecast before correctness is observable. Later outcomes grade the forecast. The question is not whether code or a model scores better, but who owns the forecast-ranking when the world cannot yet grade it. RouteCast makes the ownership boundary explicit: the model may propose, but the provisional ranking is a versioned computation. A frozen milestone predicate turns today's provisional forecast into a checkable contract at resolution time.

\subsection{Versioned Learning Loop}

Resolved outcomes do not revise the frozen forecast. They enter a separate, versioned learning loop that measures forecast error and evaluates candidate updates to priors, weights, thresholds, evidence-reweighting parameters, and calibration mappings. The objective is to improve the calibration and discrimination of future forecasts. A new scorer version is promoted only when preregistered evaluation on held-out resolved cohorts shows improvement without material integrity regressions.

\subsection{Adversarial Silence}

Internal red-teaming of the staged expected-value fold surfaced an adversarial omission pattern: deleting risky intermediate transitions from a route can raise its provisional score, because an unmodeled span pays neither cost nor probability. The same typed-transition structure that enables later resolution also makes this failure mode detectable as a deterministic coverage check over typed state records, without recourse to a model judge. A systematic treatment---including exploit measurement, a deterministic detector, and value-provenance coupling intended to close the exploit---is reserved for a companion paper.

\subsection{Domain Extension}

OSS project selection, research-direction selection, and product-roadmap prioritization are proposed transfer tests. They are not established domains. Each would need its own point-in-time packet construction, outcome mapping, leakage audit, calibration analysis, and negative-result reporting.

\section{Conclusion}

We studied code-owned provisional evaluation under delayed ground truth, a regime in which deterministic code cannot check correctness at scoring time and must instead issue a forecast-ranking. RouteCast instantiates this regime for typed strategic routes using point-in-time evidence and auditable computation. In a small retrospective pilot, the whole-packet score showed discrimination and identity exposure increased apparent LLM-judge performance, while a preregistered decomposition ablation was indistinguishable from the whole-packet protocol. Prospective transition-level resolution is required to evaluate calibration, decision utility, and cross-domain transfer.

\section*{Competing interests}
The author is developing a commercial implementation of the protocol through Dynamic Resonance.

\bibliographystyle{plain}
\bibliography{references}

\clearpage
\onecolumn
\appendix
\section{Supplementary Case-Level Pilot Table}

The full case-level pilot table is appendix-only and uses blinded IDs. Original company names and the blind-ID map are intentionally not shown in the paper.

\begin{center}
\scriptsize
\setlength{\tabcolsep}{2.5pt}
\noindent\parbox{\textwidth}{\textbf{Appendix Table A1. Case-level blinded pilot detail.} Original company names are withheld from the paper; blind IDs map to de-identified point-in-time packets. Raw scores are not cross-scorer comparable; AUC and lift use within-scorer rankings. Buckets: S+ strong-positive, + positive, \(\sim\) mixed, - negative. The RouteCast score column preserves the legacy frozen \texttt{crf\_score} field; the blind LLM column reports the mean over \(K=5\) same-model blinded passes.}
\vspace{0.5em}

\begin{tabular}{@{}p{0.8cm}p{2.55cm}p{2.15cm}p{2.05cm}p{1.75cm}p{1.7cm}p{1.75cm}p{1.05cm}@{}}
\toprule
\shortstack[l]{Blind\\ID} & Category & \shortstack[l]{RouteCast\\score \(\uparrow\)\\within-scorer\\rank only} & \shortstack[l]{Blind LLM\\mean score \(\uparrow\)\\separate scale} & \shortstack[l]{Heuristic\\score \(\uparrow\)\\separate scale} & \shortstack[l]{RouteCast\\predicted\\bucket} & \shortstack[l]{Later\\outcome\\bucket} & \shortstack[l]{Recog.\\risk} \\
\midrule
BC01 & consumer / search & 36.6 & 54 & 33 & neg & - & -- \\
BC02 & hardware / robotics & 45.4 & 58 & 69 & mix & - & -- \\
BC03 & real estate / analytics & 47.2 & 46 & 69 & mix & - & -- \\
BC04 & SaaS / hiring & 49.7 & 50 & 71 & pos & \(\sim\) & -- \\
BC05 & consumer / e-commerce & 42.5 & 57 & 67 & mix & - & -- \\
BC06 & developer tools / infrastructure & 55.6 & 60 & 89 & pos & + & yes \\
BC07 & consumer / social & 39.2 & 60 & 43 & neg & - & -- \\
BC08 & SaaS / developer API & 55.3 & 47 & 81 & pos & - & yes \\
BC09 & developer tools / cloud & 55.5 & 55 & 79 & pos & + & yes \\
BC10 & developer tools / AI API & 48.4 & 63 & 47 & mix & + & -- \\
BC11 & consumer / productivity & 31.5 & 38 & 13 & neg & - & -- \\
BC12 & SaaS / adtech & 45.0 & 52 & 67 & mix & \(\sim\) & -- \\
BC13 & SaaS / developer services & 47.6 & 52 & 59 & mix & - & -- \\
BC14 & healthcare / mental wellness & 48.2 & 54 & 45 & mix & - & -- \\
BC15 & consumer / video & 39.0 & 62 & 43 & neg & \(\sim\) & -- \\
BC16 & SaaS / web design tools & 53.0 & 57 & 67 & pos & S+ & yes \\
BC17 & consumer / on-demand services & 41.3 & 55 & 57 & neg & \(\sim\) & -- \\
BC18 & consumer / local delivery & 40.0 & 52 & 57 & neg & S+ & yes \\
BC19 & SaaS / mobile tools & 50.8 & 48 & 71 & pos & - & -- \\
BC20 & SaaS / careers & 47.4 & 61 & 67 & mix & - & -- \\
BC21 & education & 40.7 & 44 & 43 & neg & - & -- \\
BC22 & SaaS / communications API & 57.8 & 60 & 69 & pos & - & yes \\
BC23 & govtech / nonprofit software & 51.0 & 62 & 69 & pos & - & -- \\
BC24 & security / identity & 53.0 & 55 & 79 & pos & + & -- \\
BC25 & consumer / utility & 40.7 & 50 & 43 & neg & - & -- \\
BC26 & consumer / marketplace & 40.6 & 52 & 67 & neg & \(\sim\) & -- \\
\bottomrule
\end{tabular}
\end{center}

\clearpage
\section{External Evaluation Tracks}

\begin{center}
\centering
\small
\noindent\parbox{\textwidth}{\textbf{Appendix Table B1. Proposed external evaluation tracks.} These tracks are transfer tests for the prospective protocol; they do not convert the retrospective pilot into evidence of cross-domain validity.}
\vspace{0.5em}

\begin{tabularx}{\textwidth}{p{3.1cm}X X X}
\toprule
Track & Needed data owner/source & First pilot & Incentive \\
\midrule
Venture routes &
Accelerators, angel groups, venture studios, or founder communities with timestamped route packets and later outcomes. &
Blind point-in-time route ranking on an older resolved cohort, followed by a small prospective cohort. &
Better audit discipline for route selection without turning funding into the only outcome proxy. \\
\addlinespace
OSS/developer project selection &
Maintainers, developer platforms, package registries, or hackathon/program operators with public adoption traces. &
Point-in-time ranking of project directions using only repository, package, and community signals available before later adoption. &
Transparent evidence about which project bets earn durable use rather than short-lived attention. \\
\addlinespace
Novel research direction selection &
Labs, workshop organizers, OpenReview-style venues, or benchmark maintainers with timestamped proposals and later traces. &
Blind ranking of research directions before acceptance, reuse, citation, or benchmark uptake is known. &
Better evaluation of tractable novelty under noisy peer-review and reuse signals. \\
\addlinespace
Product roadmap prioritization &
Growth-stage product teams willing to share de-identified roadmap, feature, and metric histories. &
Retrospective point-in-time ranking of shipped and rejected feature bets, then preregistered prospective scoring. &
Decision support that reports uncertainty, negative evidence, and opportunity cost rather than persuasive roadmap prose. \\
\bottomrule
\end{tabularx}
\end{center}

\end{document}